\documentclass{ifacconf}
\usepackage{graphicx}
\graphicspath{{figure/}{graphics/plots/}}
\usepackage[width=1\textwidth,format=hang]{caption}
\usepackage{bm}

\usepackage{natbib}
\usepackage{tikz}
\usepackage{adjustbox}
\usepackage{pgfplots}
\usepgfplotslibrary{groupplots}
\usetikzlibrary{pgfplots.groupplots}
\usetikzlibrary{plotmarks}
\usetikzlibrary{calc}
\usepackage{comment}
\usepgfplotslibrary{external}
\pgfplotsset{compat=newest}

\usetikzlibrary{matrix}
\pgfplotsset{grid style={dashed,gray}}
\usepackage{amsmath} 
\usepackage{amssymb}  
\usepackage{color,soul}
\usepackage{bibrange}
\usepackage{hvline}
\usepackage{url}
\usepackage{multirow}
\usepackage{booktabs}
\usepackage{epstopdf}
\usepackage{subfig}

\usepackage[ruled,linesnumbered,algo2e]{algorithm2e}
\usepackage{parskip}
\usepackage{hyphenat}
\makeatletter
\usepackage{siunitx}
\usepackage{nicematrix}
\newcommand{\nosemic}{\renewcommand{\@endalgocfline}{\relax}}
\newcommand{\dosemic}{\renewcommand{\@endalgocfline}{\algocf@endline}}

\begin{document}
\begin{frontmatter}

\title{Sampling-based trajectory (re)planning for differentially flat systems: Application to a 3D gantry crane} 
\author[First]{M.N. Vu},
\author[First]{M. Schwegel},
\author[First]{C. Hartl-Nesic},
\author[First,Second]{A. Kugi}
\thanks{This work has been accepted to IFAC for publication under a Creative Commons Licence CC-BY-NC-ND.}

\address[First]{Automation and Control Institute (ACIN), 
   TU Wien, Vienna, Austria (e-mail: \{vu, schwegel, hartl, kugi\}@acin.tuwien.ac.at).}
\address[Second]{Center for Vision, Automation and Control, AIT Austrian Institute of Technology, Vienna, Austria (e-mail: \{andreas.kugi\}@ait.ac.at)}

\begin{abstract}                
In this paper, a sampling-based trajectory planning algorithm for a laboratory-scale 3D gantry crane in an environment with static obstacles and subject to bounds on the velocity and acceleration of the gantry crane system is presented.
The focus is on developing a fast motion planning algorithm for differentially flat systems, where intermediate results can be stored and reused for further tasks, such as replanning.
The proposed approach is based on the informed optimal rapidly exploring random tree algorithm (informed RRT*), which is utilized to build trajectory trees that are reused for replanning when the start and/or target states change.
In contrast to state-of-the-art approaches, the proposed motion planning algorithm incorporates a linear quadratic minimum time (LQTM) local planner.
Thus, dynamic properties such as time optimality and the smoothness of the trajectory are directly considered in the proposed algorithm.
Moreover, by integrating the branch-and-bound method to perform the pruning process on the trajectory tree, the proposed algorithm can eliminate points in the tree that do not contribute to finding better solutions. This helps to curb memory consumption and reduce the computational complexity during motion (re)planning. Simulation results for a validated mathematical model of a 3D gantry crane show the feasibility of the proposed approach. 
\end{abstract}


\end{frontmatter}
\section{Introduction}
\label{section: introduction}
Motion planning is a fundamental task in robotics that aims to at determining collision-free and dynamically feasible paths for a robotic system to reach a specified target configuration. 
Many robotic tasks require repeated planning from/to adjacent configurations, such as unloading ships, trucks, and storage.
Here, repeated movements are carried out to relocate goods that are typically packed tightly.
Hence, in many cases the motion planning problem has to be solved repeatedly for similar starting and target states.
In this paper, a motion planning algorithm based on optimal rapidly-exploring random trees (RRT*) is presented that allows for an efficient replanning for adjacent configurations.

The motion planning problem is commonly solved by discretizing the continuous state space into grids, i.e., graph-based search, or by randomly sampling the space, i.e., sampling-based search.
Graph-based search methods (deterministic motion planning searchers), see, e.g.,  A* \cite{hart1968formal}, are resolution-complete methods that guarantee optimal resolution. The search procedure is mainly guided by heuristic minimization of cost function from the current sampled state to the target state. Other versions of A* such as the Life Long Planning algorithm, see, \cite{koenig2004lifelong}, Replanning D*, see, e.g., \cite{ferguson2005delayed}, \cite{koenig2002improved}, and the Anytime algorithm ARA*, see, \cite{likhachev2003ara}, have shown that the solution can be computed and refined in a reasonable computation time depending on the chosen grid resolution. Note that while the discretized resolution is increased to obtain a better solution, the computational costs increase significantly, see, e.g., \cite{bertsekas1975convergence}. This circumstance is called the ``curse of dimensionality,'' see, \cite{ferguson2005guide}. Despite all its drawbacks, graph-based search was successfully applied to several types of planning tasks, e.g., manipulation planning, see, \cite{donald1987search}, \cite{kondo1991motion}, and kinodynamic planning, see, \cite{cherif1999kinodynamic}. 
On the other hand, sampling-based methods (probabilistic motion planning searchers) build a data tree by randomly sampling the planning space.
Some sampling-based methods have even proven to be globally probabilistically optimal.
Thus, the probability of finding the global optimal path approaches one as the number of iterations goes to infinity.
In the probabilistic roadmap (PRM), a collision-free state of a robotic system in the configuration space is selected, see, e.g, \cite{bohlin2000path}.
Then, these sampled states are connected to the respective neighboring states using a local motion planner to build the motion planning graph. The main advantage of PRM in high-dimensional configuration spaces is that only collision-free states are collected in the data tree. As a result, fewer states need to be evaluated in the search space. 

The RRT* algorithm finds the path from the initial state to each state in the planning tree by incrementally rewiring the tree of sampled states, see, e.g.,  \cite{karaman2011sampling}. Rewiring helps to reconstruct the tree by not only adding new states to the tree, but also considering them as surrogate nodes for existing states in the data tree. Further improvements of RRT* have been proposed in the literature, such as RRT*-SMART, see, e.g., \cite{nasir2013rrt}, which uses additional heuristics to speed up the convergence rate. In \cite{karaman2011anytime}, RRT* is developed and adapted for online motion planning. The robot moves along the initial path while the algorithm is still refining the part that the robot has not yet reached. In \cite{gammell2014informed} the Informed-RRT* is presented. Instead of sampling the system state in the entire workspace, this algorithm randomly samples the system state in the subspace created by the first solution. Note that there is no difference between informed-RRT* and RRT* until the first solution is found. After that, only feasible samples from the subset of states, i.e., the informed set, are allowed in the informed RRT* algorithm. This helps to narrow the search space and increase the probability of obtaining a better solution in a given time. 

Recently, \cite{strub2020advanced} introduced the advanced batch informed RRT* (ABIT*) in an effort to unify search-based and graph-based search without sacrificing the advantages of either method. This algorithm discretizes the continuous search spaces with an edge-implicit Random Geometric Graph (RGG), see \cite{karaman2011sampling}. This can improve the computation time by applying the informed RRT* in parallel for each space region created by the RGG method. A local planner computes the cost for moving the system between two sampled states. To account for the dynamic constraints of the system, a Two Point Boundary Value Problem (TP-BVP) solver, see, e.g., \cite{xie2015toward,keller2018numerical}, is utilized as local planner. \cite{webb2013kinodynamic} employed the linear quadratic minimum time (LQMT) analytical solution, see, \cite{verriest1991linear}, as local planner for RRT*. Since the LQMT solver provides an analytical solution for the near-time optimal trajectory connecting any two system states, the computation time of this local planner is minimized.  However, this local planner is only applicable for linear systems. 

Due to the high dimension of the 3D gantry crane system, i.e., $5$ degrees of freedom (DoF) corresponding to a $10$-dimensional system state, sampling-based motion planning is favorable for this system compared to grid-based motion planning. The main objective of this work is to develop a fast sampling-based trajectory planning algorithm that drives the 3D gantry crane from a given starting state to a given target state, avoiding obstacles and respecting the dynamic constraints of the system states. 
To this end, the informed RRT* algorithm is extended to meet the requirements of repeated and efficient motion (re)planning from/to adjacent configurations.

The sampling-based trajectory planning proposed in this paper, named as flat-informed RRT*, incorporates three modifications. 
    First, the informed subset of randomly sampled states is controlled using a heuristic function to prevent new sampled states that do not contribute to a better solution. 
    Second, the branch-and-bound method is used to prune the parts of the trajectory tree that do not provide a better solution compared to the current cost.
    Third, taking advantage of parallel data processing, multiple trees are generated and concatenated to provide more information within the workspace. 

To achieve a fast computation time, the local planner uses the analytical solution of LQMT. However, the 3D gantry crane has nonlinear system dynamics. To overcome this challenge, the differential flatness property of the 3D gantry crane is exploited, see, e.g., \cite{delaleau1995decoupling}, \cite{kolar2017time}, \cite{kolar2013flatness}, where all system states and inputs can be parameterized by the flat output and its time derivatives. 

Once the collision-free and dynamically feasible trajectory tree is available, the proposed algorithm quickly generates a feasible trajectory when the target state is changed.   

The paper is organized as follows. Section \ref{section: Mathematical Model} briefly introduces the mathematical model of the laboratory-scale 3D gantry crane and its differential flatness. Section \ref{section: Flat - Informed RRT*} presents the proposed RRT* motion planning algorithm with the local planner LQMT. Simulation results for a validated mathematical model of a 3D gantry crane are presented in Section \ref{section: Simulation Result} and a conclusion is drawn in Section \ref{section: Conclusion}. 

\section{Mathematical Model}
\label{section: Mathematical Model}
\begin{figure}[h!]
\centering
   \def\svgwidth{1\columnwidth}
    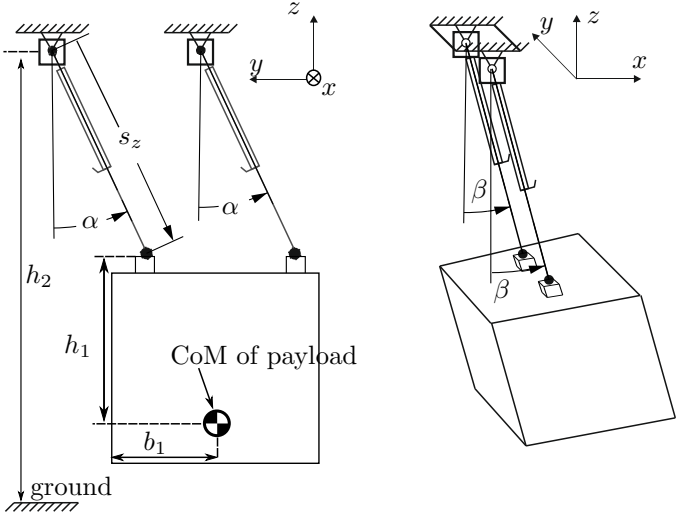
        \caption{The payload with the corresponding hosting cable angles $\alpha$ and $\beta$, see \cite{vu2022fast}.}
    \label{fig: the schematic hook model}
\end{figure}

The CAD model of the 3D gantry crane is illustrated in Fig. \ref{fig: the schematic hook model}. The gantry crane system consists of $5$ degrees of freedom $\mathbf{q} = [s_x,s_y,s_z,\alpha,\beta]^\mathrm{T}$, where $s_x$ and $s_y$ are controlled by the bridge belt driver and the trolley motor driver in $x$- and $y$-direction of the payload, respectively. To lift and lower the payload, the hoisting drum performs a movement in $z$-direction, indicated by the position of the center of mass (CoM) of the payload. The variables $\alpha$ and $\beta$ refer to the angles of the rope in the $zy$- and $zx$-plane, respectively, see Fig. \ref{fig: the schematic hook model}. The two identical ropes are modeled as massless rigid rods. Thus, the gantry crane can be treated as a rigid-body system.
\begin{figure}[h]
\centering
    \def\svgwidth{1\columnwidth}
    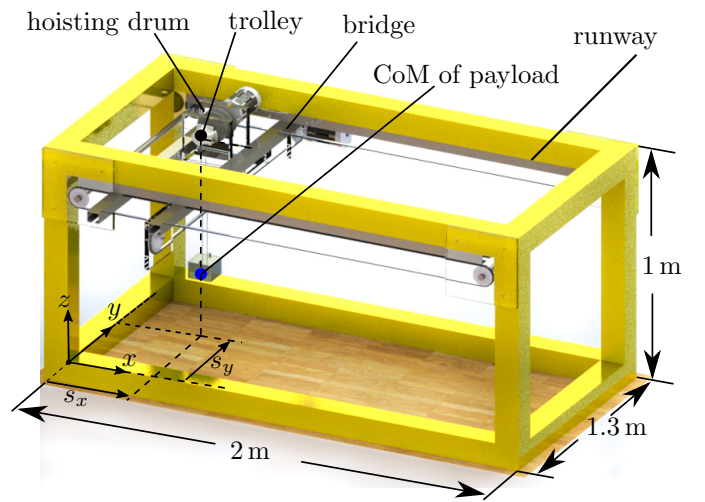
        \caption{CAD model of the  3D gantry crane \textcolor{black}{for $\alpha=\beta=0$}, see \cite{vufast}.}
    \label{fig: the schematic model}
\end{figure}
Using the five generalized coordinates $\mathbf{q}$, the state-space equations of motion of the 3D gantry crane are derived by utilizing the Euler-Lagrange equations, see \cite{lobe2018flatness}
\begin{equation}
\begin{aligned}
\dot{\mathbf{z}} = \mathbf{f}(\mathbf{z},\mathbf{u}) 
= \begin{bmatrix}
\dot{\mathbf{q}}
\\
\mathbf{M}^{-1}(\mathbf{q})\bigg( \begin{bmatrix}
\mathbf{u} \\
\mathbf{0}
\end{bmatrix} - \mathbf{C}(\mathbf{q},\dot{\mathbf{q}})\dot{\mathbf{q}} - \mathbf{g}(\mathbf{q})\bigg)
\end{bmatrix},
\end{aligned}
\label{Eq: dynamics}
\end{equation}
with the system state $\mathbf{z}^\mathrm{T} = [\mathbf{q}^\mathrm{T},\dot{\mathbf{q}}^\mathrm{T}] $. The matrix $\mathbf{M}(\mathbf{q})$ denotes the symmetric and positive definite mass matrix, $\mathbf{C}(\mathbf{q},\dot{\mathbf{q}})$ includes Coriolis and centrifugal terms, $\mathbf{g}(\mathbf{q})$ are the forces associated \textcolor{black}{with} the potential energy, and $\mathbf{u}^\mathrm{T} = [u_1,u_2,u_3] \in \mathbb{R}^3$ are the driving forces in the $x$-, $y$-, and $z$-direction, respectively.  
 
The payload is assumed to be a point mass. In this case, the gantry crane is a flat system and the position of the payload is a flat output, see, e.g., \cite{kolar2017time}, \cite{fliess1995flatness}. Considering the position of the CoM of the payload $\mathbf{p}_L = [x_L(t),y_L(t),z_L(t)]^\mathrm{T}$ as the flat output, the system state $\mathbf{z}$ and the control input $\mathbf{u}$ can be parameterized in the form
\begin{subequations}
\begin{align}
    & \mathbf{z} = \mathbf{f}_z(\mathbf{p}_L,\dot{\mathbf{p}}_L, \ddot{\mathbf{p}}_L, \mathbf{p}_L^{(3)}) \\
    & \mathbf{u} = \mathbf{f}_u(\mathbf{p}_L,\dot{\mathbf{p}}_L, \ddot{\mathbf{p}}_L,\mathbf{p}_L^{(3)}, \mathbf{p}_L^{(4)}). 
\end{align}
\label{eq: def flat system}
\end{subequations}
The flatness property is beneficial, since the proposed sampling-based trajectory planning algorithm directly provides the four-times differentiable desired trajectories  $\mathbf{p}_L^d(t)$ and the corresponding desired system states and control inputs follow from (\ref{eq: def flat system}). Furthermore, the four-times differentiable desired trajectories $\mathbf{p}_L^d$ can also be written as a the linear time-invariant system

\begin{equation}
\begin{aligned}
\dot{\mathbf{x}}_l(t) &= \mathbf{A} \mathbf{x}_l(t) + \mathbf{B}\mathbf{u}_l(t),
\end{aligned}
\label{eq: linear system}
\end{equation}
with the system matrices
\begin{equation*}
\mathbf{A} = \begin{bmatrix}
\mathbf{0} & \mathbf{I}_3 & \mathbf{0} & \mathbf{0} \\ 
\mathbf{0} & \mathbf{0} & \mathbf{I}_3 & \mathbf{0} \\
\mathbf{0} & \mathbf{0} & \mathbf{0} & \mathbf{I}_3 \\
\mathbf{0} & \mathbf{0} & \mathbf{0} & \mathbf{0}
\end{bmatrix}, \:\:\: 
\mathbf{B} = \begin{bmatrix}
\mathbf{0} \\
\mathbf{0} \\
\mathbf{0} \\
\mathbf{I}_3 \\
\end{bmatrix}, 
\end{equation*}
the state $\mathbf{x}_{l}^\mathrm{T} = [\mathbf{p}_L^d,\dot{\mathbf{p}}_L^d,\ddot{\mathbf{p}}_L^d, {\mathbf{p}_L^d}^{(3)}]$, the input $\mathbf{u}_l={\mathbf{p}_L^d}^{(4)}$, and $\mathbf{I}_n$ is the identity matrix of size $n$.
\section{Flat-Informed RRT* sampling-based trajectory planning}
\label{section: Flat - Informed RRT*}
In the first subsection, the linear quadratic local planner with minimum time is discussed, which computes the optimal trajectory between any two sampled states. Additionally, a feasibility flag for each optimal trajectory determines whether it is collision-free and dynamically feasible or not. In the second subsection, the proposed sampling-based trajectory planning is presented in detail. 
\subsection{Linear quadratic minimum time (LQMT) local planner}
\label{subsection 1: Flat-Informed RRT*}
For the sampling-based trajectory planning, the local planner is used to compute the optimal cost ${c}^*$ to move the system (\ref{eq: linear system}) from a state $\mathbf{x}_l(t_0) = \mathbf{x}_{l,0}$ to a state $\mathbf{x}_l(t_1) =\mathbf{x}_{1,1}$. This optimal cost $c^*(\mathbf{x}_{l,0},\mathbf{x}_{l,1})$ can be found by solving the linear quadratic minimum time problem, see \cite{verriest1991linear}
\begin{subequations}\label{eq: lqmt total}
    \begin{align} 
        \label{eq: lqmt main}
        \min_{\mathbf{x}_{l},\mathbf{u}_l, \Delta t} c &= \Delta t + \dfrac{1}{2}\int_{t_0}^{t_1} \mathbf{u}_l^\mathrm{T} \mathbf{R} \mathbf{u}_l\mathrm{d}t \\
        \label{eq: lqmt flat}
        \mathrm{s.t.} \: \dot{\mathbf{x}}_l(t) &= \mathbf{A}\mathbf{x}_l(t) + \mathbf{B}\mathbf{u}_l(t), \:\:\:t_0 \leq t \leq t_1  \\
        \label{eq: lqmt start}
        \mathbf{x}_l(t_0) &= \mathbf{x}_{l,0}\:\:\:\mathbf{x}_l(t_1) = \mathbf{x}_{l,1}\:\:, 
        \end{align}
\end{subequations}
where $\Delta t = t_1 - t_0$, (\ref{eq: lqmt flat}) corresponds to (\ref{eq: linear system}), and $\mathbf{R}$ is a user-defined positive definite weighting matrix that determines the tradeoff between trajectory smoothing and transit time $\Delta t$. 
The Hamiltonian $\mathcal{H}$ of (\ref{eq: lqmt total}) reads as
\begin{equation}
\mathcal{H} = 1+\dfrac{1}{2} \mathbf{u}_l^\mathrm{T} \mathbf{R} \mathbf{u}_l + \bm{\lambda}_l^\mathrm{T} (\mathbf{A}\mathbf{x}_l + \mathbf{B}\mathbf{u}_l).
\end{equation}
The necessary first-order conditions for optimality are given by
\begin{subequations}\label{eq:states costs}
\begin{align}
\label{eq:states costs 1}
& \dot{\mathbf{{x}}}_l^* = \left(\dfrac{\partial \mathcal{H}}{\partial \bm{\lambda}_l}\right)^{\!\mathrm{T}}=\mathbf{A}\mathbf{x}_l^* +\mathbf{B}\mathbf{u}_l^* \\
\label{eq:states costs 2}
& \dot{\bm{\lambda}}_l^*=-\left(\dfrac{\partial\mathcal{H}}{\partial \mathbf{x}_l}\right)^{\!\mathrm{T}} = -\mathbf{A}^{\!\mathrm{T}}\bm{\lambda}_l^* \\
\label{eq:states costs 3}
& \mathbf{0} = \left(\dfrac{\partial \mathcal{H}}{\partial \mathbf{u}_l}\right)^{\!\mathrm{T}} = \mathbf{R}\mathbf{u}_l^* + \mathbf{B}^\mathrm{T}\bm{\lambda}_l^*\:\:.
\end{align}
\end{subequations}
From (\ref{eq:states costs 2}) and (\ref{eq:states costs 3}), we get
\begin{subequations}
    \begin{align}
        \mathbf{u}_l^* &= -\mathbf{R}^{-1}\mathbf{B}^\mathrm{T}\bm{\lambda}_l^*(t)
        \label{eq: u_l optimal}\\
        \bm{\lambda}_l^*(t) &= \exp{(\mathbf{A}^{\mathrm{\!T}}(t_1-t))}\bm{\lambda}_l^*(t_1)\:\:.
        \label{eq: lambda optimal}
    \end{align}
\end{subequations}
Substituting (\ref{eq: u_l optimal}) and (\ref{eq: lambda optimal}) into (\ref{eq:states costs 1}), the optimal trajectory $\mathbf{x}_l^*$ reads as
\begin{equation}
\mathbf{x}_l^*(t) = \exp{(\mathbf{A}(t-t_0))} \mathbf{x}_{l,0} - \mathbf{G}(t_0,t)\bm{\lambda}_l^*(t_1),
\label{eq: xt}
\end{equation}
where 
\begin{equation*}
    \mathbf{G}(t_0,t) = \int_{t_0}^{t} \exp{(\mathbf{\!A}(t-\tau))}\mathbf{BR}^{-1}\mathbf{B}^\mathrm{T}\exp{(\mathbf{A}^\mathrm{T}(t_1-\tau))}\mathrm{d}\tau
\end{equation*}
is the continuous reachability Gramian. The evaluation of (\ref{eq: xt}) at $t = t_1$ yields 
\begin{equation}
    \bm{\lambda}_l^*(t_1) = -\mathbf{G}^{-1}(t_0,t_1)\mathbf{d}_{\Delta t},
    \label{eq: lambda at t1}
\end{equation}
with $\mathbf{d}_{\Delta t} = \mathbf{x}_{l,1} - \exp{(\mathbf{A}\Delta t)}\mathbf{x}_{l,0}$. By combining (\ref{eq:states costs 3}), (\ref{eq: lambda optimal}), and (\ref{eq: lambda at t1}), the optimal control input $\mathbf{u}_l^*$ is given by
\begin{equation}
\mathbf{u}_l^*(t) = \mathbf{R}^{-1}\mathbf{B}^\mathrm{T} \exp{(\mathbf{A}^{\!\mathrm{T}}(t_1-t))}\mathbf{G}^{-1}(t_0,t_1)\mathbf{d}_{\Delta t}\:\:.
\label{eq: optimal control}
\end{equation} 
Substituting the optimal control input (\ref{eq: optimal control}) into (\ref{eq: lqmt main}) yields
\begin{equation}
c(\mathbf{x}_{l,0},\mathbf{x}_{l,1}) = \Delta t + \dfrac{1}{2}\mathbf{d}_{\Delta t}^\mathrm{T} \mathbf{G}^{-1}(t_0,t_1)\mathbf{d}_{\Delta t}\:\:.
\label{eq: final J}
\end{equation}
Since the initial time $t_0$ is known, the function $c(\mathbf{x}_{l,0},\mathbf{x}_{l,1})$ depends only on the final time $t_1$. Moreover, $c(\mathbf{x}_{l,0},\mathbf{x}_{l,1})$ could be formulated as a $7$-th order polynomial of the final time $t_1$ because the system matrix $\mathbf{A}$ is a nilpotent matrix. Thereby, the optimal value of $t_1$ can be determined by finding the roots of the first order derivative of the function (\ref{eq: final J}) with respect to $t_1$. 
Subsequently, the corresponding optimal state $\mathbf{x}_{l}^*$ and the control input $\mathbf{u}_{l}^*$ are computed using (\ref{eq: u_l optimal}) and (\ref{eq: xt}). 
To check the dynamic feasibility of the optimal trajectory $\mathbf{x}_{l}^*$, ($\ref{eq: def flat system}$) is used to obtain the system state $\mathbf{z}$ and the control input $\mathbf{u}$ of the 3D gantry crane. 
Finally, the local planner LQMT provides the optimal target cost $c^*(\mathbf{x}_{l,0},\mathbf{x}_{l,1})$ along with a feasibility flag which indicates if the optimal trajectory is collision free and dynamically feasible. 
\subsection{Sampling-based trajectory planning}
\label{subsection 2: Flat - Informed RRT*}
In this subsection, the sampling-based trajectory planning is described in detail and is summarized in Algorithm 1. The algorithm is used to build a trajectory tree $\mathcal{T}$ rooted in the starting state $\mathbf{x}_{l,start}$. This trajectory tree $\mathcal{T}$ is represented as the set $\mathcal{T} = \{\mathcal{V},\mathcal{P},\mathcal{C},\mathcal{L}\}$, where $\mathcal{V} $ is the set of states $\mathbf{x}_{l}$ in the tree and the associated costs, i.e. $c^*(\mathbf{x}_{l,start},\mathbf{x}_l)$ and $c^*(\mathbf{x}_{l},\mathbf{x}_{l,target})$. The sets $\mathcal{P}$ and $\mathcal{C}$ contain the parent and child points of the corresponding states in the set $\mathcal{V}$, respectively, and the set $\mathcal{L}$ is the mask set used for the pruning process. \textcolor{black}{Note that a state ``$B$'' is considered as a child node of a state ``$A$'' if the cost ${c}^*(A,B)$  is the smallest compared to costs from the state $A$ to other nodes in the tree $\mathcal{T}$}. The cost function between any two states in the tree $\mathcal{T}$ is computed using the local planner LQMT. Important steps in the Algorithm 1 are briefly presented in the following.

First, the proposed algorithm randomly samples the flat output $\tilde{\mathbf{x}}_l$ in each iteration (line 6). A sampled trajectory point $\tilde{\mathbf{x}}_l$ in the set $\mathcal{V}$ is constructed as
\begin{equation}
\begin{aligned}
\mathcal{V}(\tilde{\mathbf{x}}_l) &= \{\tilde{\mathbf{x}}_l,c^*(\mathbf{x}_{l,start},\tilde{\mathbf{x}}_l),{c}^*(\tilde{\mathbf{x}}_l,\mathbf{x}_{l,target})\},
\end{aligned}
\end{equation}
where $c^*(\mathbf{x}_{l,start},\tilde{\mathbf{x}}_l)$ is the optimal cost used to move the gantry crane from the initial state $\mathbf{x}_{l,start}$ to the sampled state $\tilde{\mathbf{x}}_l$ through the parent state ${\mathbf{x}}_{l,parent}$. Thus, the associated cost reads as
\begin{equation*}
    c^*(\mathbf{x}_{l,start},\tilde{\mathbf{x}}_l) = c^*(\mathbf{x}_{l,start},{\mathbf{x}}_{l,parent}) + c^*({\mathbf{x}}_{l,parent},\tilde{\mathbf{x}}_{l})\:\:.
\end{equation*}

Second, the algorithm searches for the parent state $\tilde{\mathbf{x}}_{l,parent}$ of the sampled state $\tilde{\mathbf{x}}_l$ (lines $4-7$). Here, the first breadth-first search is utilized, see, e.g., \cite{kozen1992depth}. In addition, the $\mathrm{PopQueue}$ is used to obtain an item for processing while removing that item from the top of the queue. This helps to reduce the memory consumption of the algorithm and increases the computation speed. 

Third, the feasibility check is performed to verify whether the sampled state $\tilde{\mathbf{x}}_l$ can be inserted into the trajectory tree $\mathcal{T}$ or not (lines $11-15$). The feasibility check includes the two conditions
\begin{subequations}
    \begin{align}
        \label{eq: verify 1}
        &(1)\:\: \mathrm{LocalPlanner}(\mathbf{x}_{l,parent},\tilde{\mathbf{x}}_l) \in \mathbb{X}_{free} \\
        \label{eq: verify 2}
        &(2)\:\: \tilde{\mathbf{x}}_l \in \mathcal{X}_{\hat{f}_c}
    \end{align}    
\end{subequations}
where $\mathbb{X}_{free}= \mathbb{P}_{free} \times [\underline{\mathbf{z}}^\mathrm{T}, \overline{\mathbf{z}}^\mathrm{T}] \times [\underline{\mathbf{u}}^\mathrm{T}, \overline{\mathbf{u}}^\mathrm{T}]$, and the lower and upper bounds of the system state and control input in (\ref{Eq: dynamics}) are given by $\underline{\mathbf{z}}$, $\overline{\mathbf{z}}$, $\underline{\mathbf{u}}$, $\overline{\mathbf{u}}$. $\mathbb{P}_{free}$ denotes the free space not occupied by the obstacles. By substituting the result of the the local planner (\ref{eq: xt}) and (\ref{eq: u_l optimal}) into (\ref{eq: def flat system}), the 3D gantry crane system state $\mathbf{z}$ and the control input $\mathbf{u}$ can be calculated and verified in (\ref{eq: verify 1}). The second feasibility condition (\ref{eq: verify 2}) is called the ``informed set'', i.e. the set of states that provide a better solution than the current cost $J^*$ of the tree
\begin{equation}
\mathcal{X}_{\hat{f}_c}:= \bigg\{\mathbf{x}_l\in \mathbb{X}_{free}|\hat{f}_c(\mathbf{x}_l)<J^*\bigg\}\:\:,
\label{eq: informed set}
\end{equation}
with 
\begin{equation}
\hat{f}_c(\mathbf{x}_l) = c^*(\mathbf{x}_{l,start},{\mathbf{x}}_l) + {c}^*({\mathbf{x}}_l,\mathbf{x}_{l,target})\:\:.
\end{equation}
In (\ref{eq: informed set}), $J^{*}$ is the cost of moving the gantry crane from the initial state $\mathbf{x}_{l,start}$ through its intermediate child states in the tree $\mathcal{T}$ to the target state $\mathbf{x}_{l,target}$. Note that the initial value for $J^{*}$ is infinity until the proposed algorithm finds the first obstacle-free and dynamically feasible trajectory that connects $\mathbf{x}_{l,start}$ to $\mathbf{x}_{l,target}$. This total cost value of the tree $J^{*}$ is improved if the proposed algorithm finds a better solution. When the sampled state $\tilde{\mathbf{x}}_{l}$ passes the feasibility test, the trajectory tree $\mathcal{T}$ is updated, as shown in lines $16-18$ of Algorithm 1. 

Fourth, once the cost $J^*$ in the trajectory tree is found, the mask set $\mathcal{L}$ of each trajectory node $\mathbf{x}_{l}\in \mathcal{V}$ is recalculated
\begin{equation}
\mathcal{L}(\mathbf{x}_{l}) = \begin{cases}
0\:\: & \text{if}\: c^*({\mathbf{x}_{l,start},\mathbf{x}_{l}}) > J^*\\
1\:\: & \text{otherwise}\:\:.
\end{cases}
\label{eq: prunning}
\end{equation}
Using the mask set $\mathcal{L}$, the branch-and-bound technique is applied to prune the trajectory tree $\mathcal{T}$, as shown in lines $26-32$ of Algorithm 1. The advantage of this procedure is twofold. First, the update helps to limit the number of states to be checked during the initial breadth-first search (line 8). Second, any state with mask value $0$ can be eliminated since it does not contribute to finding a better solution. To enrich the trajectory tree $\mathcal{T}$, the proposed algorithm is processed in parallel to generate different trajectory tree stacks. Then, these tree stacks are concatenated into a single trajectory tree.

\setlength{\algomargin}{1.5em}
\begin{algorithm2e}
\KwIn{$(\mathbf{x}_{l,start},\mathbf{x}_{l,target}) \in \mathbb{X}_{free}$}
\KwOut{$\mathcal{T}  \leftarrow \{\mathcal{V},\mathcal{P},\mathcal{C},\mathcal{L}\}$}

\BlankLine
$\mathcal{V} \leftarrow \{\mathbf{x}_{l,start},0,{c}^*(\mathbf{x}_{l,start},\mathbf{x}_{l,target})\}$

$\mathcal{P} \leftarrow \emptyset$; $\mathcal{C} \leftarrow \emptyset$

$J^* \leftarrow \infty$; $\hat{f}_c(\mathbf{x}_{l,start}) \leftarrow 0$

\While{ $\mathsf{NotStopCriteria}$}{ 
\While{ $\mathsf{NotFeasibleSample}$}{
$\tilde{\mathbf{x}}_{l} \leftarrow \mathsf{GetRandomSample(\mathbb{X}_{free})}$

$\mathcal{Q} \leftarrow \mathbf{x}_{l,start}$

$\#$\textcolor{red}{perform the breadth-first search (BFS)}

\While{$\mathsf{isNotEmpty}(\mathcal{Q})$}{
$(\mathbf{x}_{l,min},c^*)\leftarrow \mathsf{PopQueue}(\mathcal{Q})$

\eIf{ $\textsf{LocalPlanner}(\mathbf{x}_{l,min},\tilde{\mathbf{x}}_{l}) \in \mathbb{X}_{free}$}{
 $\tilde{\mathbf{x}}_{l,parent} \leftarrow \mathbf{x}_{l,min}$}
{$\mathcal{Q} \leftarrow \mathcal{P}(\mathbf{x}_{l,min})$
} 
}
$\mathsf{NotFeasibleSample} = \tilde{\mathbf{x}}_{l} \notin \mathcal X_{\hat{f}} $

}
$\mathcal{V} \leftarrow \{\tilde{\mathbf{x}}_{l},c^*(\mathbf{x}_{l,start},\tilde{\mathbf{x}}_{l}),{c}^*(\tilde{\mathbf{x}}_{l},\mathbf{x}_{l,target})\}$

$\mathcal{P}(\tilde{\mathbf{x}}_{l}) \leftarrow \tilde{\mathbf{x}}_{l,parent}$

$\mathcal{C}(\tilde{\mathbf{x}}_{l,parent}) \leftarrow \tilde{\mathbf{x}}_{l}$

$\#$\textcolor{red}{perform the rewiring process}

\For{$\mathbf{x}_l \in \mathcal{V}$ $\mathrm{and}$ $\mathcal{L}(\mathbf{x}_{l}) = 1$}{
\eIf{$c^*(\mathbf{x}_{l,parent},\mathbf{x}_l) > c^*(\tilde{\mathbf{x}}_{l},\mathbf{x}_l)$}{
$\mathcal{P}(\mathbf{x}_l)\leftarrow {\tilde{\mathbf{x}}}_{l}$, $\mathcal{C}(\tilde{\mathbf{x}}_l)\leftarrow {\mathbf{x}}_{l}$

$\mathcal{C}(\mathbf{x}_{l,parent})\leftarrow \mathcal{C}(\mathbf{x}_{l,parent}) \setminus \mathbf{x}_l$}{continue}
}
\eIf{ $\mathsf{LocalPlanner}(\tilde{\mathbf{x}}_l,\mathbf{x}_{target})$ $\in \mathbb{X}_{free}$ }
{$J^*_{temp} \leftarrow c^*(\mathbf{x}_{l,start},\tilde{\mathbf{x}}_{l}) + c^*(\tilde{\mathbf{x}}_{l},\mathbf{x}_{l,target}) $

\eIf{$J^*_{temp} < J^*$}
{$J^*\leftarrow J^*_{temp}$\\
\:\#\textcolor{red}{pruning process}

\For{$\mathbf{x}_l \in \mathcal{V}$}{$\mathcal{L}(c^*(\mathbf{x}_{l,start},\mathbf{x}_l) > J^*) \leftarrow 0$}
}
{continue}
}{continue}
}
\caption{Flat-informed RRT* trajectory planning}
\label{alg: proposed}
\end{algorithm2e}
%

Note that the proposed algorithm can be viewed as a subclass of ``anytime'' algorithms that quickly finds a feasible (but not necessarily optimal) trajectory and gradually improves the solution over time towards global optimality, see, e.g., \cite{karaman2011anytime}. This ``anytime'' property can be achieved thanks to the structure of the trajectory tree $\mathcal{T}$, which contains extensive information about the relationship between the considered robot system and the environment. Thus, in the case of changing the target state $\mathbf{x}_{l,target}$, the refinement process that updates the associated costs on the set $\mathcal{V}$ of the trajectory tree $\mathcal{T}$ and the informed set on the mask set $\mathcal{L}$ can be performed very fast. If the trajectory tree $\mathcal{T}$ is dense enough and the deviation of the new target state from the original target state is small, the proposed algorithm can quickly find a feasible solution and gradually improves the solution if necessary. This replanning capability is advantageous because the robot system can further improve the quality of the trajectory during execution.

\section{Simulation Results}
\label{section: Simulation Result}
\begin{figure}
	\centering
	\def\svgwidth{1\columnwidth}
    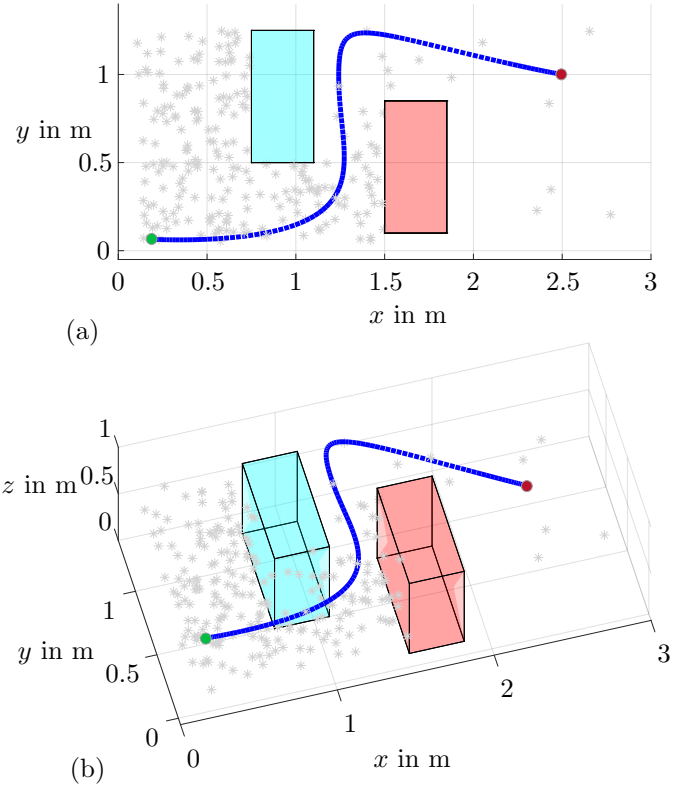
	\caption{Scenario 1: Collision-free path from a starting state (green dot) to a target state (red dot) resulting as a solution of Algorithm 1. The grey asterisks are the flat output states in the trajectory tree. (a) Path in the $xy$-plane. (b) Path in 3D space. }
	\label{fig: complex 1}
\end{figure}
\begin{figure}
	\centering
	\def\svgwidth{1\columnwidth}
    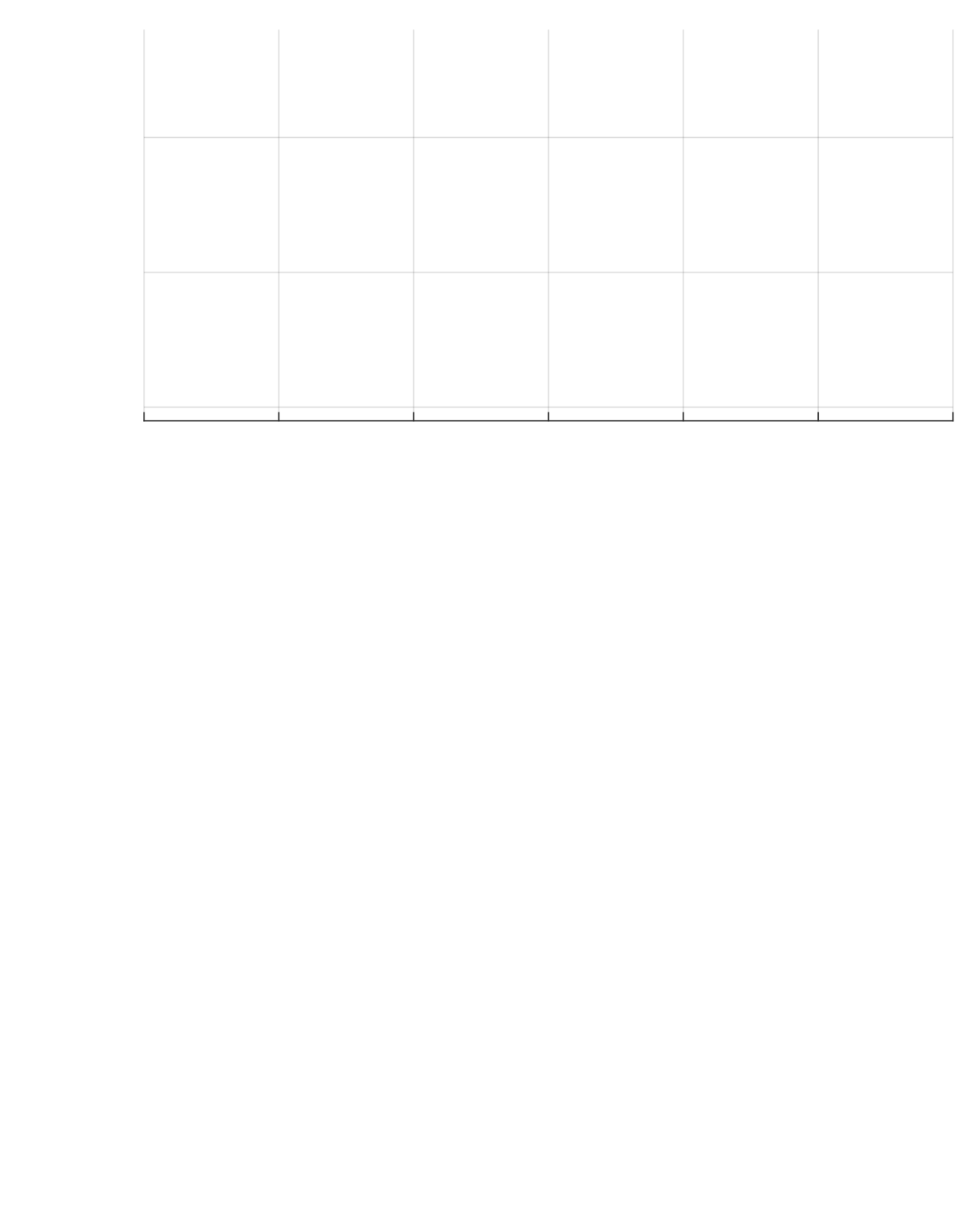
	\caption{Scenario 2: Collision-free path from a starting state (green dot) to a target state (red dot) resulting as a solution of Algorithm 1. The grey asterisks are the flat output states in the trajectory tree. (a) Path in the $xy$-plane. (b) Path in 3D space.}
	\label{fig: complex 2}
\end{figure}
\begin{figure}
    \centering
    \def\svgwidth{1\columnwidth}
%
%
\begin{tikzpicture}[line join=round]
\definecolor{acin_r}{RGB}{186,18,43}
\definecolor{acin_g}{RGB}{0,190,65}
\definecolor{acin_b}{RGB}{19,93,255}
\pgfplotsset{
	width=0.32\columnwidth,
	height=0.75in,
	at={(0.758in,3.103in)},
	scale only axis,
	yticklabel=\pgfkeys{/pgf/number format/.cd,fixed,precision=2}\pgfmathprintnumber{\tick},
	xmin=0.0000,
	xmax=15.0000,
	axis background/.style={fill=white},
	xmajorgrids,
	ymajorgrids,
	ylabsh/.style={every axis y label/.style={at={(0,0.5)}, xshift=#1, rotate=90}},
}
    \begin{groupplot}[ 
        group style={
        group size=2 by 4,
        vertical sep = 10pt,
        horizontal sep = 40pt, 
       },
       ylabsh= -3em
    ]
    \nextgroupplot[
            xticklabels={\empty},
            ylabel={$\dot{s}_x$ in $\mathrm{m}/\mathrm{s}$},
            legend style={draw=none},
            ymin= -0.6000,
			ymax= 0.6000,
    ]
    \addlegendimage{empty legend};
    \addplot [color=acin_b, line width=1pt, forget plot]
  		table[]{figure/tikz/s1_sim/flat_gc_s1-1.tsv};
  	\label{plots: desire}
	\addplot [color=black, dashed, line width=1.0pt, forget plot]
  		table[]{figure/tikz/s1_sim/flat_gc_s1-2.tsv};
	\label{plots: constrained}
	\addplot [color=black, dashed, line width=1.0pt, forget plot]
  		table[]{figure/tikz/s1_sim/flat_gc_s1-3.tsv};
	  		
    \coordinate (top) at (rel axis cs:0,1);
    \nextgroupplot[
            xticklabels={\empty},
            ylabel={$u_1$ in $\mathrm{N} \mathrm{m}$},
            legend style={draw=none},
            ymin= -0.1500,
			ymax= 0.1500,
    ]
    \addlegendimage{empty legend};
    \addplot [color=acin_b, line width=1pt, forget plot]
  		table[]{figure/tikz/s1_sim/flat_gc_s1-4.tsv};
	\addplot [color=black, dashed, line width=1.0pt, forget plot]
  		table[]{figure/tikz/s1_sim/flat_gc_s1-5.tsv};
	\addplot [color=black, dashed, line width=1.0pt, forget plot]
  		table[]{figure/tikz/s1_sim/flat_gc_s1-6.tsv};
	%
    \nextgroupplot[
            xticklabels={\empty},
			ylabel={$\dot{s}_y$ in $\mathrm{m}/\mathrm{s}$},
            legend style={draw=none},
            ymin= -0.500,
			ymax= 0.500,
    ]
	\addplot [color=acin_b, line width=1pt, forget plot]
  		table[]{figure/tikz/s1_sim/flat_gc_s1-7.tsv};
  	\addplot [color=black, dashed, line width=1.0pt, forget plot]
  		table[]{figure/tikz/s1_sim/flat_gc_s1-8.tsv};
	\addplot [color=black, dashed, line width=1.0pt, forget plot]
  		table[]{figure/tikz/s1_sim/flat_gc_s1-9.tsv};

    \nextgroupplot[
            xticklabels={\empty},
            ylabel={$u_2$ in $\mathrm{N} \mathrm{m}$},
            legend style={draw=none},
            ymin= -0.1500,
			ymax= 0.1500,
    ]
	\addplot [color= acin_b, line width=1pt, forget plot]
		table[]{figure/tikz/s1_sim/flat_gc_s1-10.tsv};
	\addplot [color=black, dashed, line width=1.0pt, forget plot]
  		table[]{figure/tikz/s1_sim/flat_gc_s1-11.tsv};
	\addplot [color=black, dashed, line width=1.0pt, forget plot]
  		table[]{figure/tikz/s1_sim/flat_gc_s1-12.tsv};

    \nextgroupplot[
    		xticklabels={\empty},
			ylabel={$\dot{s}_z$ in $\mathrm{m}/\mathrm{s}$},
            legend style={draw=none},
            ymin= -0.30,
			ymax= 0.300,
    ]
    \addplot [color=acin_b, line width=1pt, forget plot]
  		table[]{figure/tikz/s1_sim/flat_gc_s1-13.tsv};
	\addplot [color=black, dashed,line width=1pt, forget plot]
  		table[]{figure/tikz/s1_sim/flat_gc_s1-14.tsv};
	\addplot [color=black, dashed, line width=1.0pt, forget plot]
  		table[]{figure/tikz/s1_sim/flat_gc_s1-15.tsv};

    \nextgroupplot[
            xticklabels={\empty},
            ylabel={$u_3$ in $\mathrm{N}\mathrm{m}$},
            legend style={draw=none},
            ymin= -0.250,
			ymax= 0,
    ]
    \addplot [color=acin_b, line width=1pt, forget plot]
  		table[]{figure/tikz/s1_sim/flat_gc_s1-16.tsv};
	\addplot [color=black, dashed, line width=1.0pt, forget plot]
  		table[]{figure/tikz/s1_sim/flat_gc_s1-17.tsv};
	\addplot [color=black, dashed, line width=1.0pt, forget plot]
  		table[]{figure/tikz/s1_sim/flat_gc_s1-18.tsv};
    \nextgroupplot[
    		legend style={draw=none},
            legend pos=north west,
			ylabel={$\alpha$ in $^\circ$},
			ymin= -3.5,
			ymax= 3.5,
            xlabel={time in \SI{}{\second}},
            scaled y ticks=false,
    ]
	\addplot [color=acin_b, line width=1pt, forget plot]
  		table[]{figure/tikz/s1_sim/flat_gc_s1-19.tsv};
	\addplot [color=black, dashed, line width=1.0pt, forget plot]
  		table[]{figure/tikz/s1_sim/flat_gc_s1-20.tsv};
	\addplot [color=black, dashed, line width=1.0pt, forget plot]
  		table[]{figure/tikz/s1_sim/flat_gc_s1-21.tsv};
    \nextgroupplot[
    		legend style={draw=none},
            legend pos=north west,
			ylabel={$\beta$ in $^\circ$},
			ymin= -3.5,
			ymax= 3.5,
            xlabel={time in \SI{}{\second}},
    ]
	\addplot [color=acin_b, line width=1pt, forget plot]
  		table[]{figure/tikz/s1_sim/flat_gc_s1-22.tsv};
	\addplot [color=black, dashed, line width=1.0pt, forget plot]
 		table[]{figure/tikz/s1_sim/flat_gc_s1-23.tsv};
	\addplot [color=black, dashed, line width=1.0pt, forget plot]
  		table[]{figure/tikz/s1_sim/flat_gc_s1-24.tsv};

   	\coordinate (bot) at (rel axis cs:1,0);
    \end{groupplot}
	\path (top|-current bounding box.north)--
      coordinate(legendpos)
      (bot|-current bounding box.north);
	\matrix[
    matrix of nodes,
    anchor=south,
    draw,
    inner sep=0.1em,
    draw
  	]at([yshift=1ex]legendpos)
  	{
    \ref{plots: desire}& Planned trajectory&[5pt] 
    \ref{plots: constrained}& Constraint &[5pt]\\};
    
\end{tikzpicture}%
    \caption{Scenario 1: Time evolution of the states and control inputs of the system (\ref{Eq: dynamics}).}
    \label{fig: S1 time evolution}%
\end{figure}
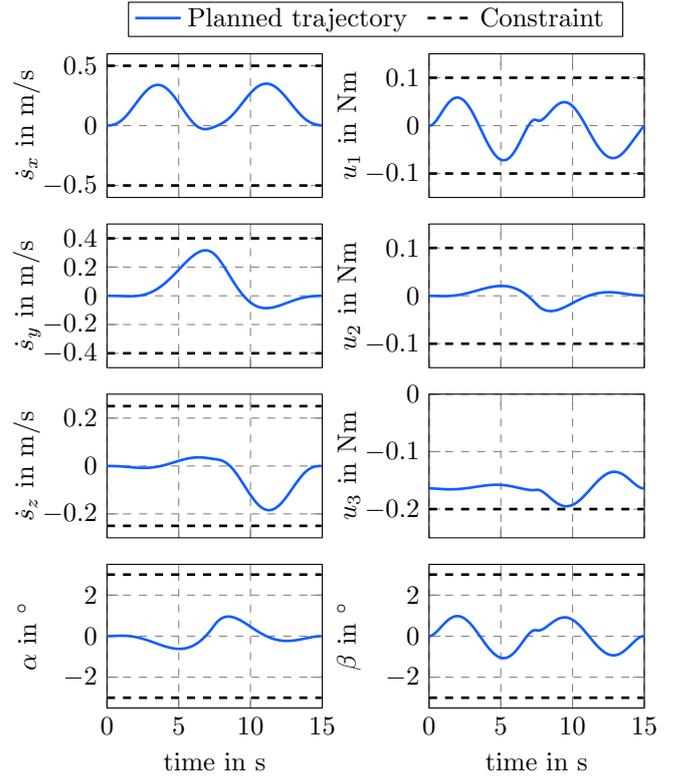
\begin{figure}
    \centering
    \def\svgwidth{1\columnwidth}
%
%
\begin{tikzpicture}[line join=round]
\definecolor{acin_r}{RGB}{186,18,43}
\definecolor{acin_g}{RGB}{0,190,65}
\definecolor{acin_b}{RGB}{19,93,255}
\pgfplotsset{
	width=0.32\columnwidth,
	height=0.75in,
	at={(0.758in,3.103in)},
	scale only axis,
	yticklabel=\pgfkeys{/pgf/number format/.cd,fixed,precision=2}\pgfmathprintnumber{\tick},
	xmin=0.0000,
	xmax=11.0000,
	axis background/.style={fill=white},
	xmajorgrids,
	ymajorgrids,
	ylabsh/.style={every axis y label/.style={at={(0,0.5)}, xshift=#1, rotate=90}},
}
    \begin{groupplot}[ 
        group style={
        group size=2 by 4,
        vertical sep = 10pt,
        horizontal sep = 40pt, 
       },
       ylabsh= -3em
    ]
    \nextgroupplot[
            xticklabels={\empty},
            ylabel={$\dot{s}_x$ in $\mathrm{m}/\mathrm{s}$},
            legend style={draw=none},
            ymin= -0.6000,
			ymax= 0.6000,
    ]
    \addlegendimage{empty legend};
    \addplot [color=acin_b, line width=1pt, forget plot]
  		table[]{figure/tikz/s2_sim/flat_gc_s2-1.tsv};
  	\label{plots: desire}
	\addplot [color=black, dashed, line width=1.0pt, forget plot]
  		table[]{figure/tikz/s2_sim/flat_gc_s2-2.tsv};
	\label{plots: constrained}
	\addplot [color=black, dashed, line width=1.0pt, forget plot]
  		table[]{figure/tikz/s2_sim/flat_gc_s2-3.tsv};
	  		
    \coordinate (top) at (rel axis cs:0,1);
    \nextgroupplot[
            xticklabels={\empty},
            ylabel={$u_1$ in $\mathrm{N} \mathrm{m}$},
            legend style={draw=none},
            ymin= -0.1500,
			ymax= 0.1500,
    ]
    \addlegendimage{empty legend};
    \addplot [color=acin_b, line width=1pt, forget plot]
  		table[]{figure/tikz/s2_sim/flat_gc_s2-4.tsv};
	\addplot [color=black, dashed, line width=1.0pt, forget plot]
  		table[]{figure/tikz/s2_sim/flat_gc_s2-5.tsv};
	\addplot [color=black, dashed, line width=1.0pt, forget plot]
  		table[]{figure/tikz/s2_sim/flat_gc_s2-6.tsv};
	%
    \nextgroupplot[
            xticklabels={\empty},
			ylabel={$\dot{s}_y$ in $\mathrm{m}/\mathrm{s}$},
            legend style={draw=none},
            ymin= -0.500,
			ymax= 0.500,
    ]
	\addplot [color=acin_b, line width=1pt, forget plot]
  		table[]{figure/tikz/s2_sim/flat_gc_s2-7.tsv};
  	\addplot [color=black, dashed, line width=1.0pt, forget plot]
  		table[]{figure/tikz/s2_sim/flat_gc_s2-8.tsv};
	\addplot [color=black, dashed, line width=1.0pt, forget plot]
  		table[]{figure/tikz/s2_sim/flat_gc_s2-9.tsv};

    \nextgroupplot[
            xticklabels={\empty},
            ylabel={$u_2$ in $\mathrm{N} \mathrm{m}$},
            legend style={draw=none},
            ymin= -0.1500,
			ymax= 0.1500,
    ]
	\addplot [color= acin_b, line width=1pt, forget plot]
		table[]{figure/tikz/s2_sim/flat_gc_s2-10.tsv};
	\addplot [color=black, dashed, line width=1.0pt, forget plot]
  		table[]{figure/tikz/s2_sim/flat_gc_s2-11.tsv};
	\addplot [color=black, dashed, line width=1.0pt, forget plot]
  		table[]{figure/tikz/s2_sim/flat_gc_s2-12.tsv};

    \nextgroupplot[
    		xticklabels={\empty},
			ylabel={$\dot{s}_z$ in $\mathrm{m}/\mathrm{s}$},
            legend style={draw=none},
            ymin= -0.30,
			ymax= 0.300,
    ]
    \addplot [color=acin_b, line width=1pt, forget plot]
  		table[]{figure/tikz/s2_sim/flat_gc_s2-13.tsv};
	\addplot [color=black, dashed,line width=1pt, forget plot]
  		table[]{figure/tikz/s2_sim/flat_gc_s2-14.tsv};
	\addplot [color=black, dashed, line width=1.0pt, forget plot]
  		table[]{figure/tikz/s2_sim/flat_gc_s2-15.tsv};

    \nextgroupplot[
            xticklabels={\empty},
            ylabel={$u_3$ in $\mathrm{N}\mathrm{m}$},
            legend style={draw=none},
            ymin= -0.250,
			ymax= 0,
    ]
    \addplot [color=acin_b, line width=1pt, forget plot]
  		table[]{figure/tikz/s2_sim/flat_gc_s2-16.tsv};
	\addplot [color=black, dashed, line width=1.0pt, forget plot]
  		table[]{figure/tikz/s2_sim/flat_gc_s2-17.tsv};
	\addplot [color=black, dashed, line width=1.0pt, forget plot]
  		table[]{figure/tikz/s2_sim/flat_gc_s2-18.tsv};
    \nextgroupplot[
    		legend style={draw=none},
            legend pos=north west,
			ylabel={$\alpha$ in $^\circ$},
			ymin= -3.5,
			ymax= 3.5,
            xlabel={time in \SI{}{\second}},
            scaled y ticks=false,
    ]
	\addplot [color=acin_b, line width=1pt, forget plot]
  		table[]{figure/tikz/s2_sim/flat_gc_s2-19.tsv};
	\addplot [color=black, dashed, line width=1.0pt, forget plot]
  		table[]{figure/tikz/s2_sim/flat_gc_s2-20.tsv};
	\addplot [color=black, dashed, line width=1.0pt, forget plot]
  		table[]{figure/tikz/s2_sim/flat_gc_s2-21.tsv};
    \nextgroupplot[
    		legend style={draw=none},
            legend pos=north west,
			ylabel={$\beta$ in $^\circ$},
			ymin= -3.5,
			ymax= 3.5,
            xlabel={time in \SI{}{\second}},
    ]
	\addplot [color=acin_b, line width=1pt, forget plot]
  		table[]{figure/tikz/s2_sim/flat_gc_s2-22.tsv};
	\addplot [color=black, dashed, line width=1.0pt, forget plot]
 		table[]{figure/tikz/s2_sim/flat_gc_s2-23.tsv};
	\addplot [color=black, dashed, line width=1.0pt, forget plot]
  		table[]{figure/tikz/s2_sim/flat_gc_s2-24.tsv};

   	\coordinate (bot) at (rel axis cs:1,0);
    \end{groupplot}
	\path (top|-current bounding box.north)--
      coordinate(legendpos)
      (bot|-current bounding box.north);
	\matrix[
    matrix of nodes,
    anchor=south,
    draw,
    inner sep=0.1em,
    draw
  	]at([yshift=1ex]legendpos)
  	{
    \ref{plots: desire}& Planned trajectory&[5pt] 
    \ref{plots: constrained}& Constraint &[5pt]\\};
    
\end{tikzpicture}%
    \caption{Scenario 2: Time evolution of the states and control inputs of of the system (\ref{Eq: dynamics}).}
    \label{fig: S2 time evolution}%
\end{figure}
\begin{figure}
\centering
	\def\svgwidth{1\columnwidth}
    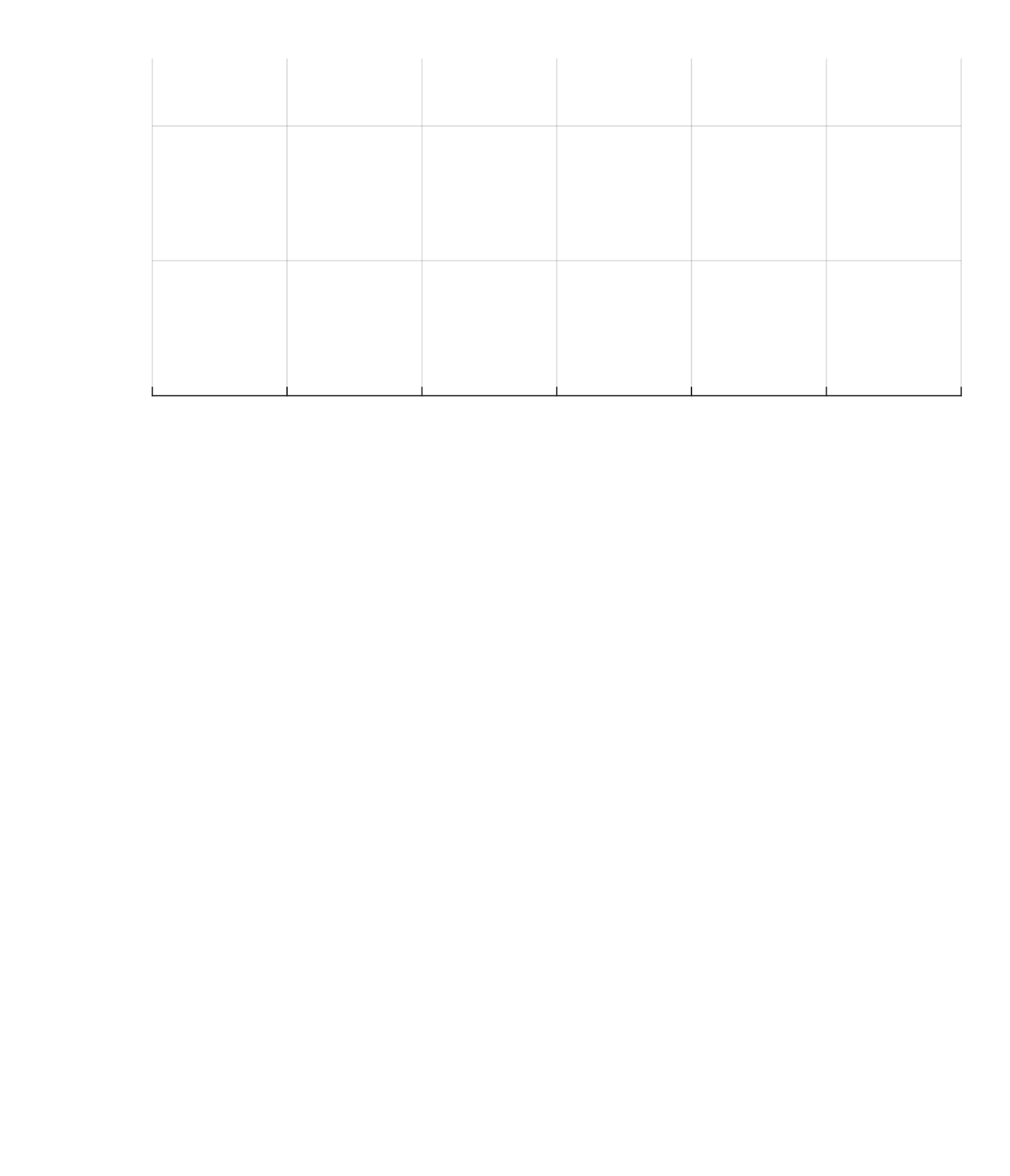
		\caption{Scenario 1: Collision-free paths (green lines) for random target payload positions. The blue path is the initial collision-free path of the tree. Red asterisks denote the random target payload positions. (a) Paths in the $xy$-plane. (b) Paths in 3D space.}
	\label{fig: complex replanning 1}
\end{figure}
\begin{figure}
\centering
	\def\svgwidth{1\columnwidth}
    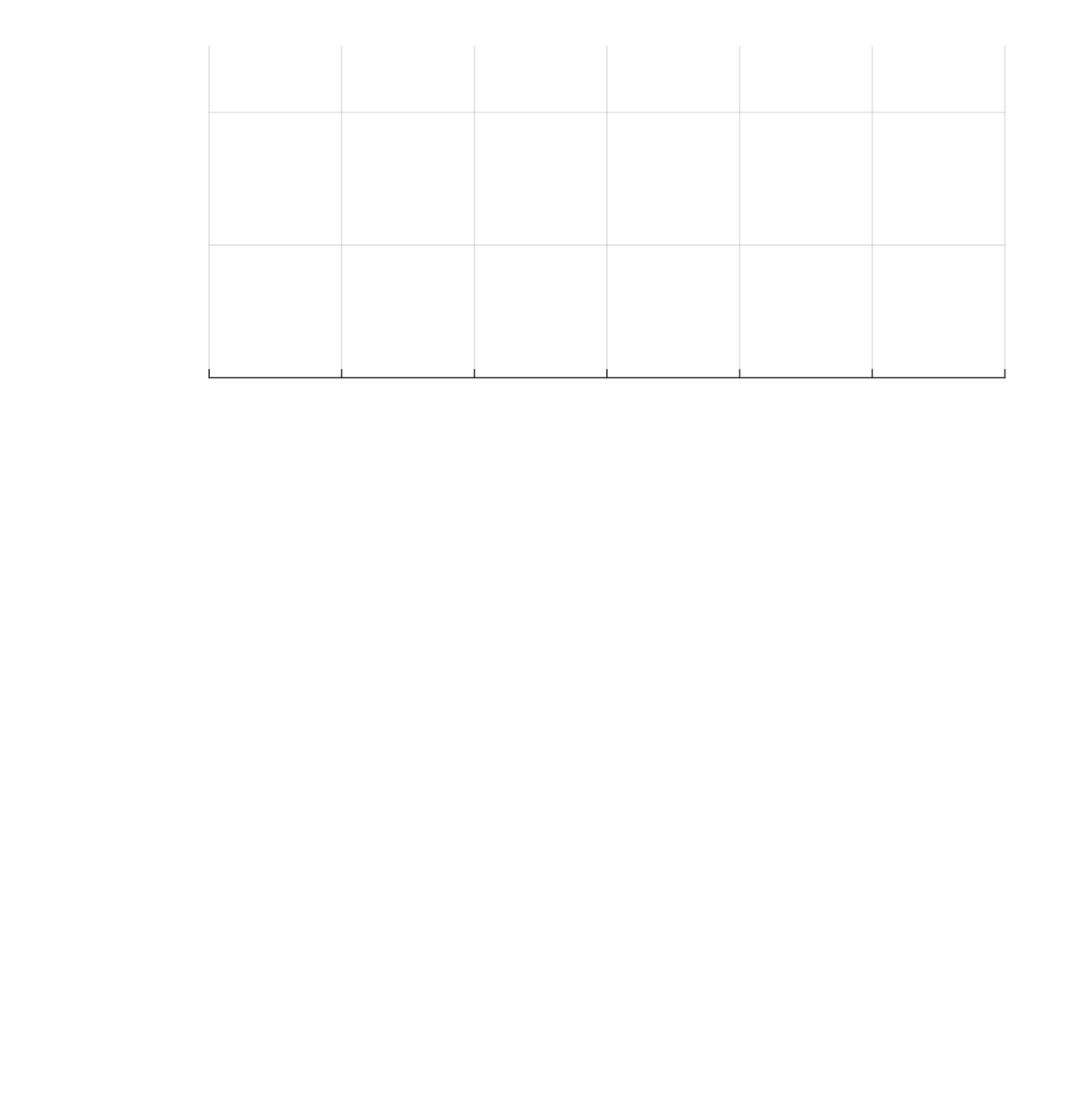
		\caption{Scenario 2: Collision-free paths (green lines) for random target payload positions. The blue path is the initial collision-free path of the tree. Red asterisks denote the random target payload positions. (a) Paths in the $xy$-plane. (b) Paths in 3D space.}
	\label{fig: complex replanning 2}
\end{figure}
The simulation was performed using \textsc{MATLAB} $\mathrm{R2021b}$ on a desktop computer with 3.4 GHz Intel Core i7 and 32GB RAM. The proposed algorithm was validated in two scenarios with different obstacle locations. For simplicity, all obstacles are on the ground and aligned with the $x$-, $y$-, and $z$-axis of the world coordinate system. The first scenario consists of two obstacles at positions $\mathbf{T}_{1} = [1.5,0.1,0]^\mathrm{T}$ and $\mathbf{T}_2 = [0.75,0.5,0]^\mathrm{T}$. \textcolor{black}{Additionally, the two boxes have the same dimension $[w,h,d]^\mathrm{T}= [0.35,0.75,0.75]^\mathrm{T}$ containing the width $w$, the height $h$, and the depth $d$}. The second scenario has three obstacles, located at the positions $\mathbf{T}_{1} = [0.735,0.20,0]^\mathrm{T}$, $\mathbf{T}_2 = [0.65,0.94,0]^\mathrm{T}$ and $\mathbf{T}_3 = [1.65,0.35,0]^\mathrm{T}$ \textcolor{black}{with the dimensions $[0.72,0.3,0.75]^\mathrm{T}$, $[0.63,0.13,0.75]^\mathrm{T}$, and $[0.35,0.45,0.75]^\mathrm{T}$, respectively}. The start and target positions of the 3D gantry crane payload in both scenarios are $\mathbf{p}_{l,start} = [0.19,0.065,0.7]^\mathrm{T}$ and $\mathbf{p}_{l,target} = [2.5,1,0.2]^\mathrm{T}$, respectively. The proposed flat-informed sampling-based trajectory planning takes random samples of the flat initial system state $\tilde{\mathbf{x}}_l$ and builds the trajectory tree according to Algorithm 1. To connect a randomly sampled system state $\tilde{\mathbf{x}}_l$ to a point in the trajectory tree $\mathcal{T}$, the local planner LQMT from Subsection \ref{subsection 1: Flat-Informed RRT*} is used with the weighting matrix $\mathbf{R} = \mathrm{diag}(0.1,0.1,0.1,0.1)$. 

In both scenarios, the planning time $t_{plan}$ of \SI{30}{\second} is used as the termination criterion, i.e. it is the maximum time allowed to build the trajectory tree. To calculate the distance from each point in the workspace to the obstacles, the two scenario maps are discretized into equidistant $3$D voxel grids of \SI{0.01}{\meter} in each dimension and the value $1$ is assigned to all voxels occupied by an obstacle. Successively, the fast Euclidean distance transform (EDT), see \cite{maurer2003linear}, is used to check whether the \textcolor{black}{trajectory of the payload} is obstacle-free or not. \textcolor{black}{Note that collisions with the ropes are not considered since the two sway angles $\alpha$ and $\beta$ are constrained in (\ref{eq: verify 1}) to a small range of $\pm 2^\circ$. This reduces the risk of rope collisions with the obstacles.}

Figs. \ref{fig: complex 1} and \ref{fig: complex 2} show collision-free paths from a payload start position (green dot) to a payload target position (red dot) in the two scenarios 1 and 2, respectively. In the first scenario, the optimal travel time $t^{*}$ is \SI{14.94}{\second}  and the optimal total cost $J^*$ is $15.32$. Although there are more obstacles in the second scenario, the optimal travel time is shorter than in the first scenario with $t^{*}=$ \SI{11.36}{\second} and the optimal total cost $J^*$ is $12.5$. Note that the density of the trajectory tree gradually decreases towards the target region because the informed property (\ref{eq: informed set}) and pruning process (\ref{eq: prunning}) of the algorithm eliminate all unnecessary trajectory points from the tree. 

Fig. \ref{fig: S1 time evolution} and \ref{fig: S2 time evolution} illustrate the time evolution of the corresponding states $\dot{s}_x$, $\dot{s}_y$, $\dot{s}_z$, $\alpha$ and $\beta$, together with the three control inputs $u_1$, $u_2$ and $u_3$ of the gantry crane system (\ref{Eq: dynamics}). Note that the state and input constraints according to (\ref{eq: verify 1}), shown as black dashed lines, are well respected. 

\captionsetup{labelsep=newline}
\begin{table}[h]
\caption{Performance on different $t_{plan}$ in $2$ scenarios}
\label{tab: MC}
\begin{center}
\begin{tabular}{c c c c c}
\textbf{Scenario 1}\\
\hline
$t_{plan}$ & \SI{30}{\second} &  \SI{50}{\second} &  \SI{100}{\second} & \SI{200}{\second}  \\
\hline
 $t^{*}$ & $14.94$ & $12.54$ & $12.84$ & $11.24$  \\
$J^{*}$ & $15.32$ & $13.96$ & $13.79$ & $12.05$  \\
$|\mathcal{T}|$ & $1102$ & $1372$ & $1979$ & $2955$ \\
\textbf{Scenario 2}\\
\hline
 $t^{*}$ & $11.36$ & $10.94$ & $9.42$ & $8.97$  \\
$J^{*}$ & $12.5$ & $11.83$ & $10.23$ & $9.75$  \\
$|\mathcal{T}|$ & $874$ & $998$ & $1049$ & $942$ \\
\hline
\end{tabular}
\end{center}

\end{table}

Table \ref{tab: MC} shows the performance of the proposed algorithm for different maximum scheduling times $t_{plan}$ for both scenarios. Note that in this simulation, the proposed algorithm is processed in parallel on $8$ CPU cores. After the maximum allowed time $t_{plan}$ elapses, all trees generated by the individual CPU cores are merged into a single tree. Overall, Table 1 shows that the travel time and the total cost decrease when the proposed algorithm is given a longer planning time $t_{plan}$. The size of the trajectory tree is denoted by $|\mathcal{T}|$. In the classical RRT* algorithm, increasing the planning time leads to a more complex data tree. However, the proposed algorithm uses the informed set (\ref{eq: informed set}) and performs the pruning procedure (\ref{eq: prunning}), which reduces the complexity of the trajectory tree. For example, in Table \ref{tab: MC} of scenario 2, although the planning time $t_{plan}$ is increased from \SI{100}{\second} to \SI{200}{\second}, the size of the trajectory tree $|\mathcal{T}|$ is reduced from $1049$ nodes to $942$ nodes. This also helps to increase the computational speed compared to the classical RRT* algorithm. 


In both scenarios, the proposed algorithm not only provides a smooth flat output trajectory that can be used to parameterize the system state using (\ref{eq: def flat system}), but also makes available the trajectory tree $\mathcal{T}$ that can be reused in the replanning process when the target state is changed. If a new target state is obtained, the trajectory tree $\mathcal{T}$ is updated by recalculating the cost of the trajectory points in the set $\mathcal{V}$ with respect to the new target state. In case the current trajectory tree can generate a collision-free and dynamically feasible trajectory to the new target state, the remainder of the tree $\mathcal{T}$ remains unchanged. Otherwise, Algorithm 1 is processed again with the current trajectory tree as the initial tree. If the trajectory tree is dense enough and the new target state does not have a large deviation from the current target state, the replanning process is very fast. This property is presented in the following simulations. 

Figs. \ref{fig: complex replanning 1} and \ref{fig: complex replanning 2} show the replanned trajectories from the initial payload position (green dot) to random target payload positions (red asterisks) for two scenarios.
The blue trajectories are the initial obstacle-free trajectories and the green trajectories are the replanned trajectories to multiple random target payload positions.
Note that the target payload positions are randomly generated with a maximum deviation of \SI{0.3}{\meter} from the current target payload position in each $x$-, $y$-, and $z$-direction.
In this simulation, the trajectory tree does not need to be recalculated because the tree $\mathcal{T}$ is dense and all generated paths are collision-free and smooth in both scenarios. 
However, even small deviations can lead to drastically different paths, due to the dynamic constraints of the crane and the presence of obstacles.
As is shown in this example, the necessary information to deal with these nonlinear changes in the global optimization is captured by the tree.
The average computation time of the rescheduling process in scenarios 1 and 2 is \SI{70}{\milli\second} and \SI{50}{\milli\second}, respectively. 
\section{Conclusion}
\label{section: Conclusion}
This paper presents the sampling-based flat-informed RRT* trajectory planning for the 3D gantry crane in an environment with static obstacles. By considering an informed set, the randomly sampled flat initial state is only considered if it leads to an improvement in the solution of the global planning problem. This reduces the computation time of the algorithm. Also, the branch-and-bound technique is employed to prune the tree and reduce the complexity of the data tree structure. The analytical solution of the linear quadratic minimum time (LQMT) local planner is used to speed up the computation time of the local planner, which is repeatedly processed within the sampling-based trajectory algorithm. Furthermore, the trajectory tree structure can be quickly updated in case the target state changes. This is helpful in practical scenarios with repetitive tasks, such as picking up and placing goods in a factory or port, where the trajectory tree can be re-used instead of solving the optimization problem repeatedly.  The proposed trajectory planning method is well suited for systems with flat outputs, such as gantry cranes and unmanned aerial vehicles (UAVs). 

Future work will focus on developing the trajectory tracking controller framework and implementing the proposed trajectory planning concept in experimental setups. 
\bibliography{MOTIONPLANNING2019BIB}

%


\end{document}